\documentclass[10pt,twocolumn,letterpaper]{article}

\usepackage[pagenumbers]{iccv} 

\definecolor{iccvblue}{rgb}{0.21,0.49,0.74}
\usepackage[pagebackref,breaklinks,colorlinks,allcolors=iccvblue]{hyperref}
\usepackage{cuted}
\usepackage{multirow}
\usepackage[accsupp]{axessibility} 
\usepackage{adjustbox}
\usepackage{soul}

\def\paperID{10043} 
\def\confName{ICCV}
\def\confYear{2025}

\title{DreamID: High-Fidelity and Fast diffusion-based Face Swapping via Triplet ID Group Learning}

\author{Fulong Ye\footnotemark[1], 
Miao Hua\footnotemark[1], Pengze Zhang, Xinghui Li, Qichao Sun, Songtao Zhao\footnotemark[2], Qian He\footnotemark[2], Xinglong Wu\\
Intelligent Creation Team, ByteDance\\
\url{https://superhero-7.github.io/DreamID/}
}

\begin{document}
\maketitle
\renewcommand{\thefootnote}{\fnsymbol{footnote}} 
\footnotetext[1]{Equal contribution.}
\footnotetext[2]{Corresponding Author.}
\begin{strip}
    \vspace*{-15mm}
    \centering
    \includegraphics[width=1\textwidth]{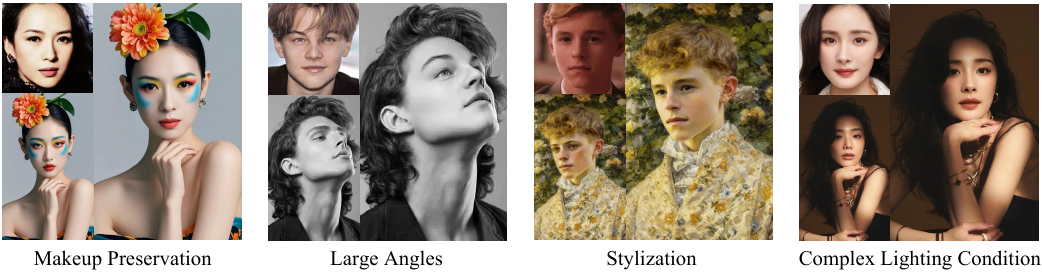}
    \captionof{figure}{DreamID can generate high fidelity face swapping results at 512 × 512 resolution. In each group, we present the swapped face on the right, which is created by replacing the source face (top-left) with the target face (bottom-left). Our model is capable of generating high-similarity face swaps and performs exceptionally well in a variety of challenging scenarios, including makeup preservation, large angles, stylization, and complex lighting conditions.}
    \label{teaser}
\end{strip}
\begin{abstract}
In this paper, we introduce \textbf{DreamID}, a diffusion-based face swapping model that achieves high levels of ID similarity, attribute preservation, image fidelity, and fast inference speed. Unlike the typical face swapping training process, which
often relies on implicit supervision and struggles to achieve satisfactory results. DreamID establishes explicit supervision for face swapping by constructing Triplet ID Group data, significantly enhancing identity similarity and attribute preservation. The iterative nature of diffusion models poses challenges for utilizing efficient image-space loss functions, as performing time-consuming multi-step sampling to obtain the generated image during training is impractical. To address this issue,  we leverage the accelerated diffusion model SD Turbo, reducing the inference steps to a single iteration, enabling efficient pixel-level end-to-end training with explicit Triplet ID Group supervision. Additionally, we propose an improved diffusion-based model architecture comprising SwapNet, FaceNet, and ID Adapter. This robust architecture fully unlocks the power of the Triplet ID Group explicit supervision. Finally, to further extend our method, we explicitly modify the Triplet ID Group data during training to fine-tune and preserve specific attributes, such as glasses and face shape. Extensive experiments demonstrate that DreamID outperforms state-of-the-art methods in terms of identity similarity, pose and expression preservation, and image fidelity. Overall, DreamID achieves high-quality face swapping results at 512×512 resolution in just 0.6 seconds and performs exceptionally well in challenging scenarios such as complex lighting, large angles, and occlusions.
\end{abstract}

\section{Introduction}
Face swapping is a highly challenging task that aims to transfer identity-related information from a source image to a target image while preserving the attribute information of the target image, such as background, lighting, expression, head pose, and makeup.

\begin{figure}[htbp] 
\centering 
\includegraphics[width=1.0\columnwidth]{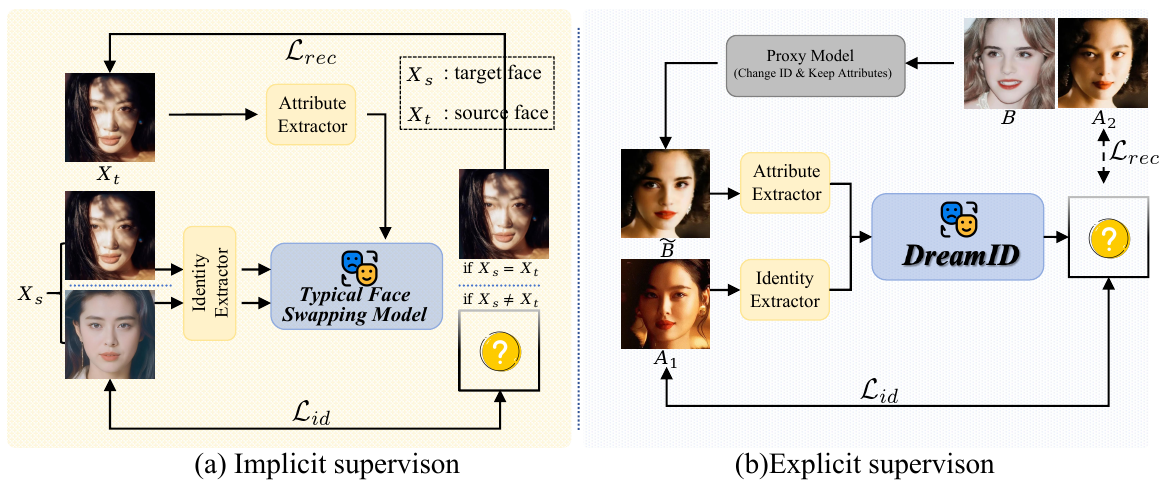} 
\caption{(a) The typical face-swapping training process, which often relies on implicit supervision. (b) Unlike previous work, DreamID constructs Triplet ID Group data for explicit supervision.}
\label{fig: intro}
\end{figure}

Early face swapping research primarily focused on GAN-based methods \cite{simswap,simswapplusplus,e4s,cscs}, which encountered two primary challenges. First, the training process tends to be unstable and requires extensive hyperparameter search. Second, the generated images often suffer from low fidelity and various artifacts, particularly in scenarios with large angles and the edges of the face shape. Addressing these challenges within the GAN framework is particularly difficult. Recently, diffusion models \cite{stable_diffusion,dalle2,imagen,altdiffusion,anydressing} have achieved remarkable success in image generation, demonstrating significant advantages in image fidelity and diversity. Based on this, recent studies \cite{DiffFace,diffswap,FaceAdapter,ReFace} introduce the diffusion model to the face swapping task. While these approaches have significantly improved the quality of image generation, they still fail to achieve satisfactory face swapping results. The primary challenge in face swapping lies in the absence of real ground truth, that is, it is difficult to find a “real” swapped image for a given $\{source\  image\ X_s, target\  image \ X_t\}$ pair. As illustrated in Figure~\ref{fig: intro}(a), previous works primarily inject ID and attribute information into the diffusion model through implicit supervision via an ID loss with the source image(when $X_s\neq X_t$) and a reconstruction loss with the target image(when $X_s = X_t$). Due to the lack of explicit supervision, they struggle to achieve high ID similarity and retain fine-grained attribute details such as lighting and makeup.

To overcome these limitations, this paper proposes an accurate and explicit supervised training framework for the face swapping task by constructing \textbf{Triplet ID Group} data to enhance both ID similarity and attribute retention capabilities of face swapping models. Specifically, as shown in Figure \ref{fig: intro}(b), we prepare two images sharing the same ID ($A_1$, $A_2$), and a third image has a different ID ($B$). We use a GAN-based proxy face swapping model to swap the $A_2$'s face on $B$ and get a pseudo target image $\tilde{B}$. The Triplet ID Group is (source $A_1$, pseudo target $\tilde{B}$, ground truth $A_2$). Here, $A_2$ has the same ID information as $A_1$ and the same attribute information as $\tilde{B}$. This makes $A_2$ an ideal target for face swapping. Specifically, when $A_1$ serves as the source and $\tilde{B}$ serves as the target, the theoretical face swapping Ground Truth is $A_2$. After constructing the Triplet ID Group data, it is necessary to find an appropriate loss function for end-to-end training. The iterative nature of diffusion models poses challenges for utilizing various practical image-space loss functions, such as ID loss and reconstruction loss. Specifically, incorporating these losses requires accumulating gradients across multiple denoising steps during training, which is computationally expensive. To address this issue, we leverage the recent accelerated diffusion model SD Turbo \cite{sdturbo}, which reduces the inference steps to just one. This allows us to employ image space loss functions efficiently and significantly improve inference speed.

Additionally, we propose an improved diffusion model architecture for face swapping. Our model architechture is composed of three components:1) the base Unet which we refer to as SwapNet, is responsible for the main process of face swapping. 2) the face Unet feature encoder, named FaceNet, which extracts pixel-level ID information of the user image. 3) the ID Adapter that extracts the semantic-level ID information of the user image. Finally, to further extend our method, we explicitly modify the Triplet ID Group data during training to fine-tune and preserve specific attributes, such as glasses and face shape. 

We conduct extensive testing on the FFHQ \cite{stylegan} dataset, and the results indicate that our method outperforms previous state-of-the-art (SOTA) methods in terms of ID similarity, pose, expression preservation, and image quality from quantitative metrics. Moreover, qualitative results show that our method performs exceptionally well in challenging face swapping scenarios such as occlusion, complex lighting and large angles. Overall, due to stronger supervision and a more robust model structure, our model significantly surpasses previous methods and can generate 512×512 fidelity face swapping results with high similarity and attribute preservation, while requiring only one step in inference that takes $0.6$ seconds. In summary, our contributions are as follows:
\begin{itemize}
\item{} We propose a novel precise and explicit supervision face swapping training framework by constructing Triplet ID Group data. In this framework, we employ accelerated diffusion models to reduce the number of inference steps to 1. This enables the effective utilization of image-space loss functions for end-to-end training with the Triplet ID Group. As a result, we significantly improve ID similarity and attribute preservation, while also substantially enhancing inference speed.
\item{} We introduce an improved diffusion-based model architecture comprising three modules: SwapNet, FaceNet, and ID Adapter. This robust architecture fully unlocks the power of the Triplet ID Group explicit supervision.
\item{} Extensive experimental results demonstrate that our DreamID significantly outperforms previous methods in terms of ID similarity, attribute preservation, and image fidelity.
\end{itemize}

\section{Related Work}
\subsection{GAN based Face Swapping Model}

Due to the powerful generative capabilities of GANs \cite{NIPS2014_gan}, they are widely used for face swapping. DeepFakes \cite{DBLP:journals/corr/abs-2005-05535} was specifically trained to swap faces between paired identities but is limited to those identities. Research following the disentanglement paradigm for subject-agnostic face swapping includes FaceShifter \cite{li2020advancing}, which adaptively integrates identity and attribute embeddings attentively. HifiFace \cite{conf/ijcai/0002CZCTWLWHJ21} utilizes 3D shape-aware identity features to enhance the quality of swapped faces. SimSwap \cite{simswap} introduced Weak Feature Matching Loss to improve attribute preservation, while SimSwap++ \cite{simswapplusplus} enhanced model efficiency. FaceDancer \cite{facedancer} introduced an adaptive feature fusion attention (AFFA) module to fuse attribute features adaptively. CSCS \cite{cscs} and ReliableSwap \cite{ReliableSwap} developed a reverse pseudo-input generation approach for additional training data. Recently, StyleGAN-based models \cite{stylegan,stylegan2} have been adopted for high-resolution face swapping, with MegaFS \cite{megafs} pioneering the use of StyleGAN2 \cite{stylegan2} as a decoder. InfoSwap \cite{infoswap} proposed an identity contrastive loss that better disentangles the StyleGAN latent space. RAFSwap \cite{RAFSwap} introduced a Region-Aware Face Swapping network for identity-consistent, high-resolution swapping in a local-global manner. StyleSwap \cite{styleswap} designed a novel Swapping-Guided ID Inversion strategy to improve identity similarity. Despite these advances, such methods are prone to artifacts, particularly under large pose variations and occlusions \cite{Kammoun2022GenerativeAN}. These challenges highlight the need for a closer examination of alternative approaches.

\subsection{Diffusion based Face Swapping Model}

Compared to GANs, the training of diffusion models is more stable and can generate higher quality images. Recently, some works have started to explore the use of diffusion models for face swapping. DiffFace \cite{DiffFace} is the first to use a diffusion model for face swapping. It trains an ID-Conditional DDPM \cite{ddpm}, samples with facial guidance, and uses a target-preserving blending strategy. However, as they use ID features only to train the model and the swapping is entirely done at the inference stage, this approach significantly burdens the inference process, leading to low throughput even at lower resolutions. DiffSwap \cite{diffswap} reformulates face swapping as a conditional inpainting task guided by identity features and facial landmarks. To introduce identity constraints during training, they propose a midpoint estimation method that can generate swapped faces in only 2 steps. However, the method struggles to achieve effective ID transferability due to the blurry inference results. FaceAdapter \cite{FaceAdapter} uniformly models the face swapping and face reenactment tasks. However, it does not perform well on a single task. ReFace \cite{ReFace} frames the face-swapping problem as a self-supervised, train-time inpainting task. Because the face area of the target image is masked off during the inference process, it becomes difficult to preserve the attributes. In our method, we construct Triplet ID Group data to establish precise and explicit supervision, thereby boosting ID similarity and attribute preservation.

\section{Preliminaries}

\textbf{Stable Diffusion Turbo} In this paper, we employ Stable Diffusion Turbo(SD Turbo) \cite{sdturbo} as our base model.SD Turbo is a distilled version of Stable Diffusion \cite{stable_diffusion}, which allows sampling large-scale foundational image diffusion models in 1 to 4 steps at high image quality. SD Turbo is a latent diffusion model that operates in the latent space of an autoencoder $\mathcal{D}(\mathcal{E}(\cdot))$, where $\mathcal{E}$ and $\mathcal{D}$ represent the encoder and decoder, respectively. For a given image $\textbf{x}_0$ with its corresponding latent feature $\textbf{z}_0=\mathcal{E}(\textbf{x}_0)$, the diffusion forward process is defined as:
\begin{equation} \label{eauation: noise}
    \textbf{z}_t = \sqrt{\alpha_t} \textbf{z}_0 + \sqrt{(1-\alpha_t} \epsilon, 
\end{equation}
where $\alpha_t = \prod_{s=1}^t(1-\beta_s)$, $\epsilon \sim \mathcal{N}(0, 1)$, and $\beta_s$ is the pre-defined variance schedule at timestep $s$. 
\section{Method}

\begin{figure*}[htbp] 
\centering 
\includegraphics[width=0.95\textwidth]{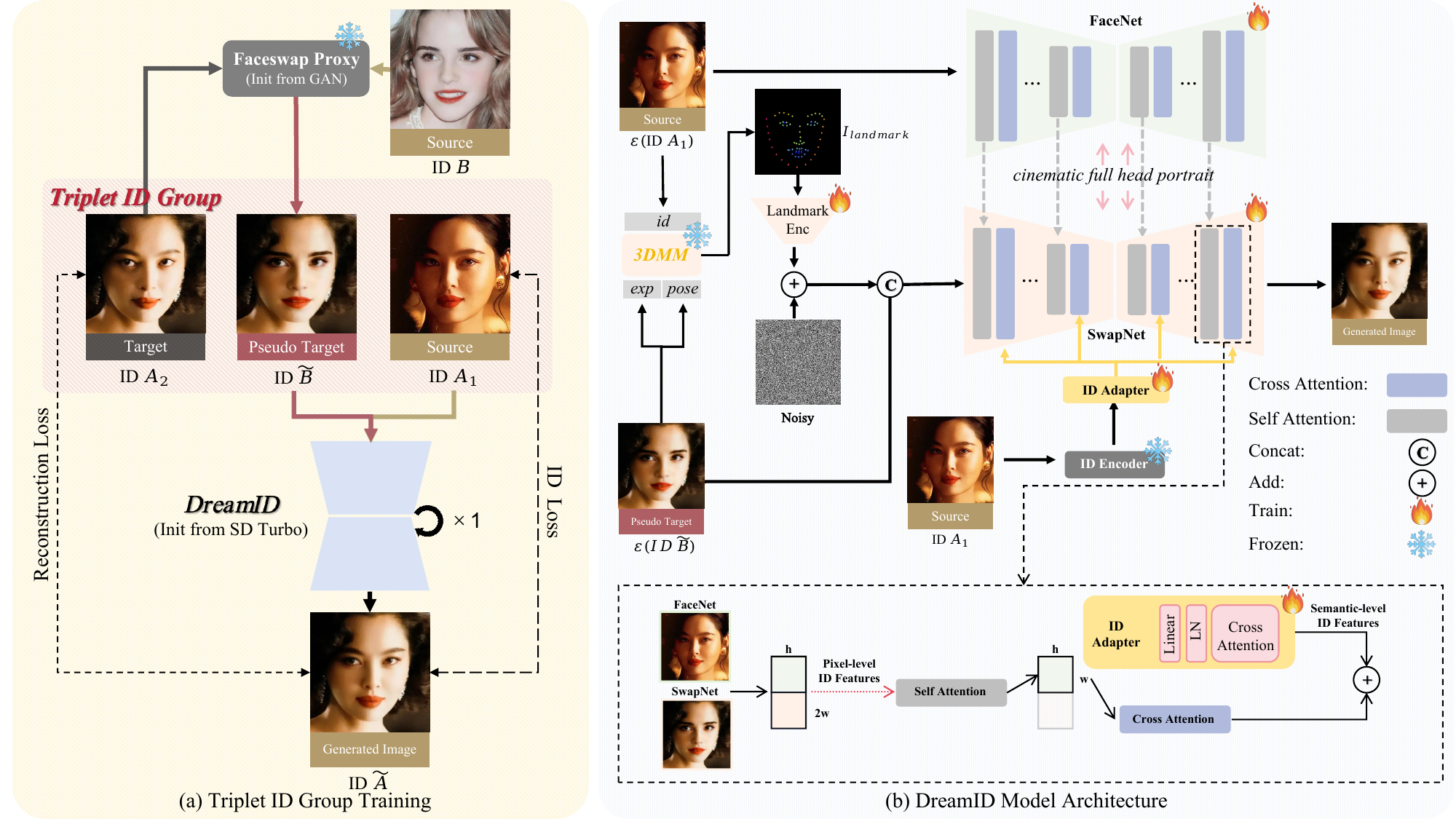} 
\caption{Overview of DreamID. (a)Triplet ID Group Training. We establish explicit supervision for face swapping by constructing Triplet ID Group data. The construction process utilizes two images with the same $\mathrm{ID}(A_1 ,A_2)$ and one image with a different $\mathrm{ID}_{B}$, along with a FaceSwap Proxy model, to generate a Pseudo Target $\mathrm{ID}_{\Tilde{B}}$. Additionally, we initialize our DreamID with SD Turbo, reducing the inference steps to a single step. This allows for convenient computation of image-space losses, such as ID Loss and reconstruction Loss. (b) DreamID Model architecture. Our model architecture is composed of three components:1) The base Unet, which we refer to as SwapNet, is responsible for the main process of face swapping. 2) the face Unet feature encoder, named FaceNet, which extracts pixel-level ID information of the user image. 3) the ID Adapter that extracts the semantic-level ID information of the user image. The core feature fusion computation process is illustrated at the bottom.}
\label{fig: model.pdf}
\vspace*{-4mm}
\end{figure*}

This chapter will provide a detailed introduction to our diffusion-based face swapping method, called \textbf{DreamID}. In Section~\ref{method:framework}, we present our precise and explicit supervision training framework. We construct Triplet ID Group data to perform pixel-to-pixel level training, significantly improving ID similarity and attribute preservation. Section~\ref{method:model_architechture} specifically details our improved diffusion-based face swapping model architecture. In section~\ref{method:finetune}, we discuss further extensions for our DreamID. An overview of our method is illustrated in Figure~\ref{fig: model.pdf}.

\subsection{Triplet ID Group Learning\label{method:framework}}

The face swapping task can be formally defined as follows:

\[ y_{\text{res}} = F(x_{\text{src}}, x_{\text{tar}}) \]
\[ \{I|y_{\text{res}}\} = \{I_{\text{id}}|x_{\text{src}}\} \cup \{I_{\text{nid}}|x_{\text{tar}}\} \]

\noindent where \( x_{\text{src}} \) is the source image, \( x_{\text{tar}} \) is the target image, \( y_{\text{res}} \) is the face swapping result image, and \( F \) denotes the face swapping model function. The objective of face swapping is to seamlessly integrate the identity information $I_{\text{id}}$ from the source image with the non-identity attribute information $I_{\text{nid}}$ of the target image (e.g., background, expression, head pose, lighting, etc.). The primary challenge in face swapping is the lack of real ground truth. Unlike typical data-driven tasks that use pair data for explicit end-to-end training, finding a ``real'' swapped image for a given $\{x_{src},x_{tar}\}$ pair is difficult. As depicted in Figure~\ref{fig: intro}, previous methods rely on implicit supervision via an ID loss with the source image and a reconstruction loss with the target image. However, this implicit supervision is biased and can lead to improper convergence. To address this, we introduce a novel Triplet ID Group Learning strategy to establish precise and explicit supervision.\\
\noindent \textbf{Triplet ID Group Construction} As illustrated in Figure~\ref{fig: model.pdf}(a), given two images of the same identity $\mathrm{ID}_{A_{1}}$ and $\mathrm{ID}_{A_{2}}$, and another image with a different identity $\mathrm{ID}_{B}$. We use $\mathrm{ID}_B$ and $\mathrm{ID}_{A_{2}}$ as the source image and target image inputs for a GAN face swapping proxy model. Then we generate a Pseudo Target $\mathrm{ID}_{\Tilde{B}}$ that alters the ID information of $\mathrm{ID}_{A_{2}}$ while preserving the attribute information. Thus, we can obtain \textbf{\textit{Triplet ID Group}} as follows:
\begin{equation}
\left(\mathrm{ID}_{A_{1}}, \mathrm{ID}_{\Tilde{B}},  \mathrm{ID}_{A_{2}}\right)
\end{equation}

Consequently, $\mathrm{ID}_{A_{2}}$ shares the same attribute information as $\mathrm{ID}_{\Tilde{B}}$ and the same identity information as $\mathrm{ID}_{A_{1}}$. This means, we have found the paired data for face swapping: when $\mathrm{ID}_{A_{1}}$ is employed as the source and $\mathrm{ID}_{\Tilde{B}}$ is used as the target, then the theoretical Ground Truth is $\mathrm{ID}_{A_{2}}$. Thus, we can use the Triplet ID Group data to perform end-to-end training. Notably, the Pseudo Target $\mathrm{ID}_{\Tilde{B}}$ generated by GAN face swapping proxy model is not our learning target; instead, our learning target is the real image $\mathrm{ID}_{A_{2}}$. This ensures that the upper bound of the supervisory signal is very high.

\noindent \textbf{Triplet ID Group Training Objectives} After constructing the Triplet ID Group data, it is necessary to find an appropriate loss function. However, applying effective image-space loss function, such as ID loss and Reconstruction Loss in diffusion models is non-trivial, due to the iterative denoising nature of diffusion models, incorporating these loss functions requires accumulating gradients across multiple denoising steps during training, which is computationally expensive. To takle this issue, we leverage recent fast sampling methods SD Turbo~\cite{sdturbo} to reduce the number of inference steps to one. Consequently, we can efficiently employ image space loss functions and significantly improve inference speed. Ultimately, we use three loss functions as our training objectives, i.e. the original Diffusion Loss, the ID Loss and the Reconstruction Loss. Next we will introduce the three loss function respectively.

\noindent \textit{Diffusion Loss} Let $x=\mathrm{ID}_{A_{2}}\in R^{3 \times H \times W}$ in the image space, $\mathcal{E}$ as the encoder module in the SD Turbo, $\mathbf{C}$ as condition encoder(details in Section~\ref{method:model_architechture}). The diffusion loss \cite{stable_diffusion} function $\mathcal{L}_{DM}$ is defined as follows:
\begin{equation} \label{equation: SD}
    \mathcal{L}_{DM} = \mathbb{E}_{\mathbf{z}_0, \epsilon, \mathbf{c}, t} \Vert \epsilon - \epsilon_\theta(\mathbf{z}_t, \mathbf{c}, t) \Vert_2. 
\end{equation}
where, $z_0=\mathcal{E}(x)\in R^{3 \times H^{'} \times W^{'}}$ in the latent space, $\epsilon \sim \mathcal{N}(0, \mathbf{I})$. And $z_t$ get from Equation~\ref{eauation: noise}, $\theta$ is the parameter of UNet to predict the noisy $\epsilon_\theta\left(z_t, c, t\right)$ condition on $z_t$, source $\mathrm{ID}_{A_{1}}$ and target $\mathrm{ID}_{\Tilde{B}}$ condition embedding $\mathbf{c}=\mathbf{C}(\mathrm{ID}_{A_{1}}, \mathrm{ID}_{\Tilde{B}})$ and $t$. Here we use one step property of SD Turbo for diffusion loss calculating, which means $t=999$. 

\noindent \textit{ID and Reconstruction Loss} Futhermore, we use one step property of SD Turbo to conduct ID and Reconstruct Loss. As shown in Figure~\ref{fig: model.pdf}, $\mathrm{ID}_{A_{1}}$ and $\mathrm{ID}_{\Tilde{B}}$ are used as the source and target input of DreamID respectively, and generated image $\mathrm{ID}_{\Tilde{A}}$ is obtained through one-step inference. Then we calculate ID loss between $\mathrm{ID}_{\Tilde{A}}$ and $\mathrm{ID}_{A_{1}}$:
$L_{id}=1-\cos \left(e_{\mathrm{ID}_{A_{1}}}, e_{\mathrm{ID}_{\Tilde{A}}}\right),$
where $e_{\mathrm{ID}_{A_{1}}}, e_{\mathrm{ID}_{\Tilde{A}}}$ present ID Embedding obtained from an off the shelf ID encoder \cite{glint36k}. Then we calculate L2 loss between $\mathrm{ID}_{\Tilde{A}}$ and $\mathrm{ID}_{A_{2}}$:
$L_{\text {rec}}=\left\|\mathrm{ID}_{A_{2}}-\mathrm{ID}_{\Tilde{A}}\right\|_2^2 .$
Finally, we get the total loss:
\begin{equation}
\mathcal{L}=\lambda_{id} \mathcal{L}_{id}+\lambda_{DM} \mathcal{L}_{DM} + \lambda_{rec} \mathcal{L}_{rec}
\end{equation}
where, $\lambda_{id},\lambda_{DM},\lambda_{rec}$ adjust the weights between each loss.

\subsection{Model Architecture\label{method:model_architechture}}

After establishing the Triplet ID Group Learning framework, the remaining task is to construct a diffusion model structure that is strong enough to fully leverage this robust supervisory signal. In this section, we will discuss our improved diffusion-based model architecture in detail. Our model architecture is composed of three components:1) the base Unet which we refer as SwapNet, is responsible for the main process of face swapping. 2) the face Unet feature encoder, named FaceNet, which extracts pixel-level ID information of the user image. 3) the ID Adapter that extracts the semantic-level ID information of the user image. As shown in Figure~\ref{fig: model.pdf}(b), we will provide detailed explanation of each component as follows. 

\noindent \textbf{SwapNet} For the base UNet module, we initialize it from SD-Turbo, thereby enabling single-step inference and significantly enhancing the inference speed. As of the input for our base UNet, we concatenate two components as follows: 1) The Facial landmarks, i.e. $I_{landmark}$. We employ a 3D face reconstruction Model~\cite{3dmm} to separately extract the identity, expression, and pose coefficients of the source and target image. Subsequently, we recombine the identity coefficients of the source image with the expression and pose coefficients of the target image to reconstruct a new 3D face model, and project it to obtain the corresponding facial landmarks. During training, we directly use the pseudo target's landmarks and feed them into a Pose Guider to extract features. This Pose Guider uses simple convolutional layers to align the landmark image with the same resolution as the noise latent. Finally, we add this feature to the noise together. 2) The latent of target image, i.e., $\mathcal{E}(\mathrm{ID}_{\Tilde{B}})$. Directly inputting the latent of the target image into the base UNet allows the model to directly capture its attribute information. Combined with the Triplet ID Group data, this enables the model to learn attribute preservation in an end-to-end manner. Finally, we expand the convolutional layer of UNet to 8 channels initialized with zero weights.

\noindent \textbf{FaceNet}  We employ a UNet encoder, namely the FaceNet encoder, which inherits weights from the SwapNet module, to encode source images.  Given the latent of a source image $\mathcal{E}(\mathrm{ID}_{A_{1}})$,we pass it through the UNet encoder to obtain a latent intermediate feature, which is then concatenated with the latent intermediate feature from the SwapNet. As shown at the bottom of Figure~\ref{fig: model.pdf}(b), we compute the self-attention on the concatenated features and then pass only the first-half dimensions from SwapNet(similar to \cite{animate_anyone}). FaceNet inputs the original user image directly, enabling strong feature extraction directly from the image space. We refer to the extracted features as pixel-level ID features.

\noindent \textbf{ID Adapter} While FaceNet is capable of extracting strong pixel-level ID features, its strength can sometimes lead to issues such as ``copy-paste'' artifacts. To balance this, we introduce the ID Adapter, which complements FaceNet by extracting semantic-level ID features. We first use a face ID encoder model to extract face embeddings, then map these embeddings to the same dimensions as the key-value (KV) matrix inputs of the SwapNet using a Liner layer and LayerNorm(LN). We then augment the SwapNet KV matrix with an extra copy of these mapped face embeddings (similar to \cite{ip_adapter}). Finally, the cross-attention from the ID Adapter and the cross-attention from SwapNet are add together. 

During training, we freeze the ID encoder and the 3DMM model\cite{3dmm}, while training all other components, including the SwapNet, FaceNet, and ID Adapter. The text prompt ``cinematic full head portrait'' is used for all training samples. After being encoded by OpenCLIP-VIT/H\cite{openclip}, the text prompt is injected into SwapNet and FaceNet through cross-attention.

\subsection{Extention\label{method:finetune}}

\begin{figure}[htbp] 
\centering 
\includegraphics[width=0.95\columnwidth]{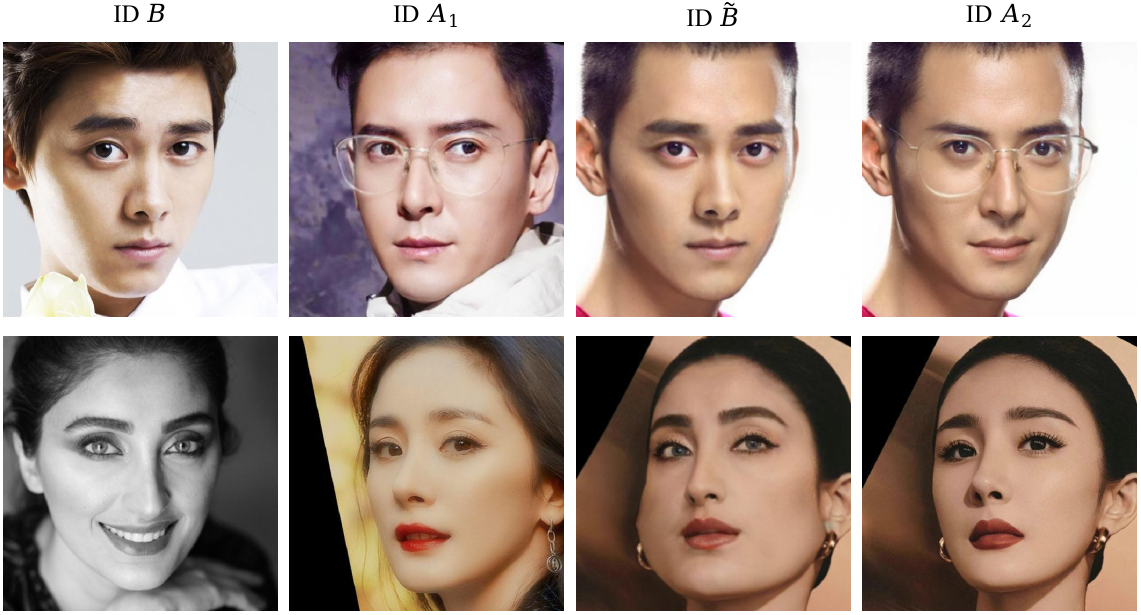} 
\caption{Data construction for specific feature control.}
\label{fig: specific_feature_tuning}
\vspace*{-4mm}
\end{figure}

In our approach, we can control specific features by explicitly modifying the Triplet ID Group data, which enables various real-world applications. For illustration, we provide two examples involving face glasses and shape transfer. As shown in Figure~\ref{fig: specific_feature_tuning}, if we want to keep the glasses from the user image, we first filter out the data with the same glasses for $\mathrm{ID}_{A_{1}}$ and $\mathrm{ID}_{A_{2}}$, and then use the post-processing model \cite{take_off_glass} to remove the glasses of the pesudo target. This enforces the model to preserve the glasses attribute from the user image. Similarly, for face shape transfer, we can employ a face shape post-processing model \cite{tps} to alter the Pseudo Target's face shape, thereby promoting the transfer of face shape from the user image.

\section{Experiment}

\begin{figure*}[htbp] 
\centering 
\includegraphics[width=0.97\textwidth]{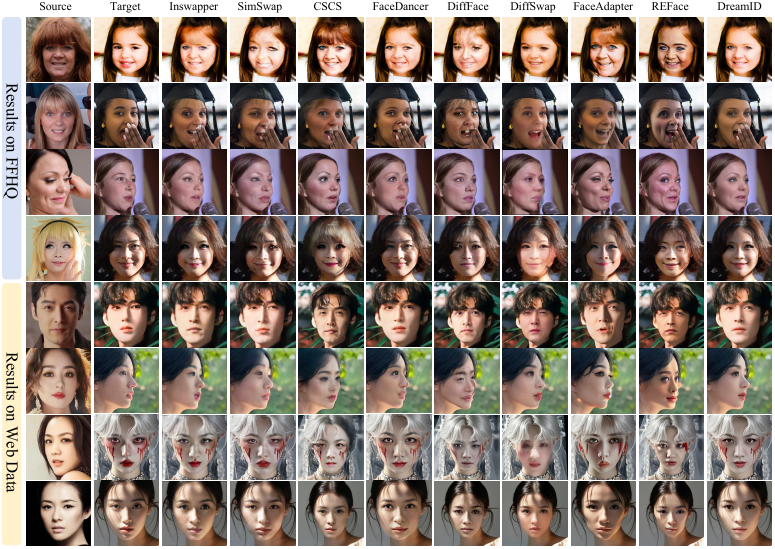} 
\caption{Qualitative comparison of state-of-the-art methods on the FFHQ dataset. DreamID demonstrates significant advantages in terms of similarity, natural blending, occlusion handling, and attribute preservation such as expression, lighting, and makeup.}
\label{fig: main_results}
\vspace*{-4mm}
\end{figure*}

\subsection{Setup}

\noindent\textbf{Datasets and Metrics} Our training data is sourced from VGGFace2-HQ \cite{simswapplusplus} and Arc2Face \cite{arc2face}, with both datasets containing multiple images per ID. We filtered these images based on clarity and similarity, ultimately selecting approximately 500,000 high-quality samples. We followed prior work \cite{cscs}, obtaining 1,000 source and 1,000 target images from FFHQ \cite{stylegan}, and generating 1,000 swapped images. We evaluated fidelity by calculating the FID between the face-swapped images and the real images from the FFHQ dataset. For pose and expression evaluation, we utilized HopeNet \cite{hope_net} and Deep3DFaceRecon \cite{deep3dface}, respectively, and compared the target and swapped images using L2 distance. We used ArcFace \cite{arcface} to extract ID embeddings and evaluated similarity by calculating the cosine distance between the ID embeddings of the face-swapped images and the source images. Additionally, we performed face retrieval by searching for the most similar faces among all the source images using cosine similarity as the measurement, and calculated the Top-1 and Top-5 accuracy. Both training and testing were conducted at a resolution of 512×512.

\noindent\textbf{Implementation Details}
We implemented our model using PyTorch, and all experiments were conducted on 8 NVIDIA A100 GPUs (80GB). Starting from SD Turbo, we trained the model with a learning rate of 1e-5 and a batch size of 8. The training process encompassed 70,000 steps, taking approximately three days. The weights for ID loss, diffusion loss, and reconstruction loss were set to 1, 1, and 10, respectively. For ID embedding and ID loss computation, we used Glint36k \cite{glint36k} as our ID encoder.

\subsection{Main Results}

\noindent\textbf{Quantitative Evaluation} In Table~\ref{Tab: main_results}, we compare with SoTA methods quantitatively on FFHQ test set, including Inswapper \cite{inswapper}, SimSwap \cite{simswap}, CSCS \cite{cscs}, DiffFace \cite{DiffFace}, DiffSwap \cite{diffswap}, FaceAdapter \cite{FaceAdapter}, REFace \cite{ReFace}. DreamID outperforms previous methods across all metrics. It achieves an FID score of 4.69, indicating that our model generates higher-fidelity images. In terms of ID similarity, our model achieves a score of 0.71, demonstrating that DreamID can effectively transfer ID information. Furthermore, DreamID achieves a Pose score of 2.20 and an Expression score of 0.789, indicating its excellent ability to preserve attributes.
\setlength{\tabcolsep}{0.3mm}{
\begin{table}[htbp]
\centering
\resizebox{\columnwidth}{!}{
\begin{tabular}{c|cccccc}
\hline
Model       & FID$\downarrow$   & \begin{tabular}[c]{@{}c@{}}ID \\ Similarity$\uparrow$\end{tabular} & \begin{tabular}[c]{@{}c@{}}ID Retrieval\\ (Top-1/Top-5)$\uparrow$\end{tabular} & Pose$\downarrow$ & Expression$\downarrow$ \\ \hline
Inswapper\cite{inswapper}   & 8.03 & 0.65 & \underline{99.20\%/99.90\%}  & 2.74 & 1.51      \\
SimSwap\cite{simswap} & 19.77 & 0.55 & 95.24\%/97.09\%  & 3.21 & 1.742      \\
CSCS\cite{cscs} & 10.17 & \underline{0.68} & 99.10\%/99.80\%  & 3.81 & 1.493      \\
FaceDancer\cite{facedancer} & \underline{4.91} & 0.48 & 92.70\%/97.20\%  & \underline{2.32} & \underline{0.854}      \\
DiffFace\cite{DiffFace} & 8.66  & 0.51 & 93.40\%/97.60\%  & 3.78 & 1.280      \\
DiffSwap\cite{diffswap} & 8.65  & 0.32 & 66.50\%/46.17\%  & 2.84 & 1.084      \\
FaceAdapter\cite{FaceAdapter} & 9.39  & 0.52 & 93.50\%/97.60\%  & 4.15 & 1.188      \\
REFace\cite{ReFace} & 5.58  & 0.57 & 96.50\%/99.20\%  & 3.77 & 1.040      \\ \hline
DreamID & \textbf{4.69}  & \textbf{0.71} & \textbf{99.9\%/100\%} & \textbf{2.20} & \textbf{0.789}      \\ \hline
\end{tabular}
}
\caption{Quantitative compare with SOTAs on the FFHQ. DreamID outperforms previous methods in all metrics, demonstrating its superiority in fidelity, ID similarity and attribute preservation.}
\label{Tab: main_results}
\vspace*{-4mm}
\end{table}

\begin{table}[htbp]
\centering
\resizebox{\columnwidth}{!}{
\begin{tabular}{cccccc}
\hline
Method          & DiffFace\cite{DiffFace} & DiffSwap\cite{diffswap} & Face Adapter\cite{FaceAdapter} & REFace\cite{ReFace} & DreamID  \\ \hline
Single Inference Speed & 25.8s      & 7.82s     & 3.42s         & 3.75        & \textbf{0.6s} \\ \hline
\end{tabular}
}
\caption{Comparison of inference time of diffusion based models.}
\label{Tab: main_results}
\vspace*{-4mm}
\end{table}

\begin{table}[htbp]
\centering
\resizebox{\columnwidth}{!}{
\begin{tabular}{lcccc}
\hline
\multicolumn{1}{c}{Model} & FID$\downarrow$  & ID Similarity$\uparrow$ & Pose$\downarrow$ & Expression$\downarrow$ \\ \hline
Full Model w/o IDA                    & 5.59 & 0.70 & 2.66 & 0.867      \\
Full Model w/o FaceNet           & 3.69 & 0.63 & 2.09 & 0.740      \\
Full Model w/o ID Loss                & \textbf{3.13} & 0.32 & \textbf{1.63} & \textbf{0.619}      \\ \hline
Full Model           & 4.69 & \textbf{0.71} & 2.20 & 0.789      \\ \hline
\end{tabular}
}
\caption{Ablation study about model architecture and training strategy.}
\label{Tab: ablation}
\vspace*{-4mm}
\end{table}

\noindent\textbf{Qualitative Evaluation} As shown in Figure~\ref{fig: main_results}, we performed qualitative comparisons with SoTAs on the FFHQ and Web data. DreamID demonstrates significant advantages in terms of similarity, natural blending, occlusion handling, and attribute preservation such as expression, lighting, and makeup. As shown in row 1, nearly all other models introduce artifacts in the bangs area of the hair and cause significant interference with expressions. In contrast, our results blend very naturally without introducing any noise, and the expressions are well-preserved. For occlusion handling, as seen in row 2, almost all other models fail to preserve occlusions effectively, whereas our model nearly perfectly maintains the occluded parts. Our method performs exceptionally well in preserving various fine-grained properties of the target image. For example, gaze in row 3, lighting in rows 4 and 8, and makeup in row 7. Even with large side profiles, our model still generates excellent results, which is a significant challenge for previous face-swapping models, where almost all other methods fail to produce reasonable outcomes. Overall, our model significantly outperforms others in terms of similarity, which is particularly noticeable in the fifth row.

\begin{figure*}[htbp] 
\centering 
\includegraphics[width=0.97\textwidth]{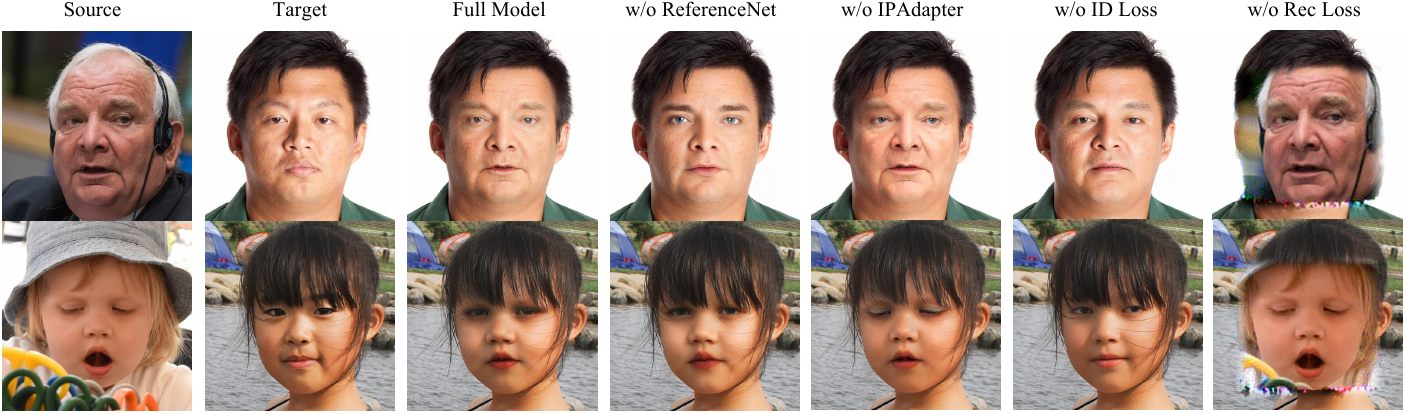} 
\caption{Visualization of ablation studies of model architecture and training strategy.}
\label{fig: ablation}
\vspace*{-4mm}
\end{figure*}
}

\begin{figure*}[htbp] 
\centering 
\includegraphics[width=0.98\textwidth]{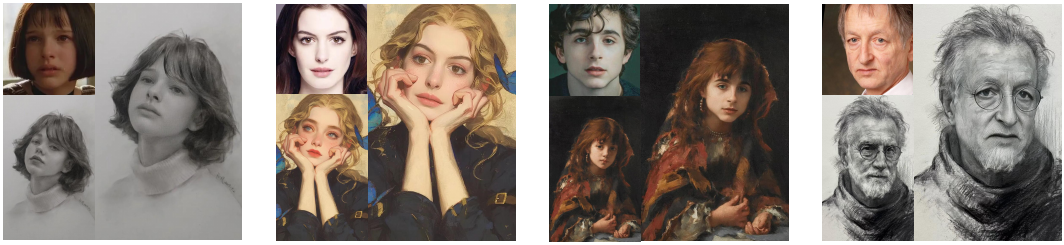} 
\caption{Results on out of domain data. DreamID effectively preserves the texture features of the template image, enabling it to generate high-quality results even in styles outside the domain of real human.}
\label{fig: style}
\vspace*{-4mm}
\end{figure*}

\noindent\textbf{Inference Speed} We test the inference speed on a NVIDIA A100. Our method only takes 0.6s for a single inference, which is much faster than other diffusion based methods.

\subsection{Ablation\label{exp:ablation}}

We conducted ablation studies on the model architecture and training strategies.

\noindent\textbf{Ablations on Model Architecture} As shown in the Table~\ref{Tab: ablation}, ``Full Model'' represents the model with both FaceNet and ID Aadapter module. ``w/o IDA'' represents the model without ID Adapter module, and ``w/o FaceNet'' represents the model without FaceNet module. Removing ID Adapter will slightly reduce ID similarity, and both Pose and Expression will also decrease. Removing FaceNet will significantly reduce ID similarity but increase Pose and Expression. By visualizing the observations in the Figure~\ref{fig: ablation}, the FaceNet is more capable of extracting pixel-level ID information, but it is more likely to have ``copy paste'' pattern. In contrast, ID Adapter is relatively weaker in feature extraction but excels at extracting semantic-level ID information. By combining the two, we can leverage the strong feature extraction capabilities of FaceNet while using ID Adapter to encourage the extraction of more essential ID features, thereby preventing the extraction of non-ID information such as expressions.

\noindent\textbf{Ablations on Training Strategies} As shown in Figure~\ref{fig: ablation}, removing the Reconstruction Loss (Rec Loss) results in unreasonable outputs. This is related to the inherent strength of diffusion models, which have a tendency to easily perform copy-pasting. Therefore, it is crucial to balance ID loss and Rec Loss against each other. If ID loss is removed, the similarity significantly decreases, as shown in Table~\ref{Tab: ablation}, indicating the importance of ID loss. In summary, both losses are complementary and indispensable.

\begin{table}[htbp]
\centering
\resizebox{\columnwidth}{!}{
\begin{tabular}{lcccc}
\hline
\multicolumn{1}{c}{Model} & FID$\downarrow$  & ID Similarity$\uparrow$ & Pose$\downarrow$ & Expression$\downarrow$ \\ \hline
DreamID(w Inswapper)                & 5.89 & \textbf{0.72} & 2.90 & 0.975 \\
DreamID(w FaceDancer)          & \textbf{4.69} & 0.71 & \textbf{2.20} & \textbf{0.789}      \\\hline
\end{tabular}
}
\vspace*{-2mm}
\caption{Ablation on proxy model selection.}
\label{Tab: ablation_proxy_model}
\vspace*{-4mm}
\end{table}

\noindent\textbf{Ablations on Proxy Model} 
In our Triplet ID Group training framework, we discovered an intriguing property: ID similarity largely depends on the ID loss between the generated image $\mathrm{ID}_{\Tilde{A}}$ and the source image $\mathrm{ID}_{A_{1}}$, while attribute preservation largely depends on the reconstruction loss between the $\mathrm{ID}_{\Tilde{A}}$ and $\mathrm{ID}_{A_{2}}$. This means that the better the Pseudo Target retains the attributes of $\mathrm{ID}_{A_{2}}$, the better the attribute preservation capability of the trained model after learning through the reconstruction loss. In other words, we need a proxy model that excels in attribute preservation, even if its ID transfer capability is not as critical. With this in mind, we compared the effects of using two different models as the proxy model. FaceDancer has good attribute retention but low ID similarity, while Inswapper is the opposite. As shown in Table~\ref{Tab: ablation_proxy_model}, the FID, Pose, and Expression of the model trained with FaceDancer are significantly better than those trained with Inswapper, while only slightly reducing the ID similarity by about 1 percentage point(statistically insignificant). Therefore, we ultimately chose FaceDancer as our proxy model. Another interesting observation is that DreamID's pose/expression scores (2.20/0.789) are better than those of FaceDancer (2.32/0.854). This indicates that the Triplet ID Group training, with real images serving as ground truth, enables the trained model to break through the upper bound of attribute preservation of the proxy model.

\begin{figure}[htbp] 
\centering 
\includegraphics[width=0.98\columnwidth]{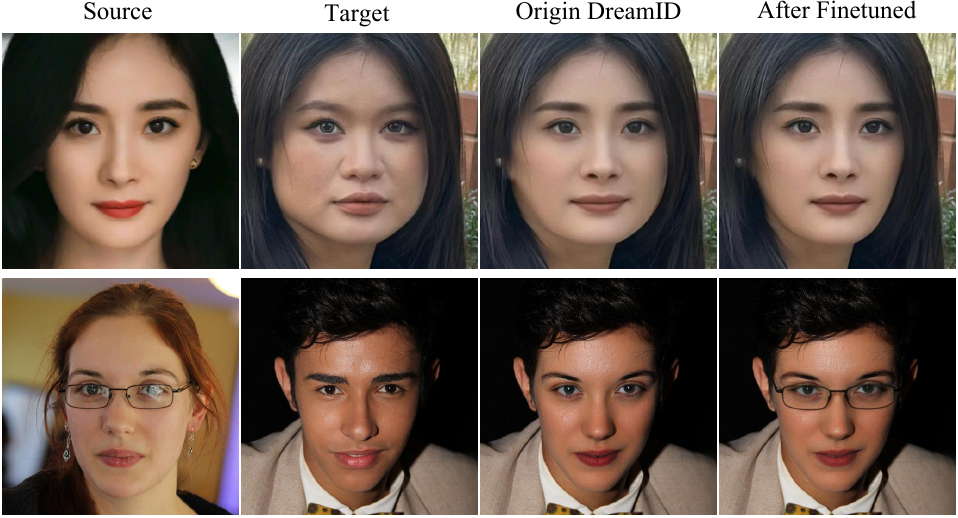} 
\caption{Illustrate results of Feature-Specific Control Finetune.}
\label{fig: faceshape_and_glass}
\vspace*{-4mm}
\end{figure}

\noindent\textbf{Feature-Specific Control Finetune} As shown in Figure~\ref{fig: faceshape_and_glass}, after fine-tuning the model as mentioned in Section~\ref{method:finetune}, both face shape and glasses can be transferred.

\noindent\textbf{Out of Domain Results} One property of DreamID is its ability to effectively preserve the texture features of the target image, which enables it to generate high-quality results in styles outside the real human domain, such as sketches, oil paintings, watercolors, etc., as shown in Figure~\ref{fig: style}.

\section{Conclusion}

In this paper, we introduced DreamID, a diffusion-based face swapping method that achieves high identity similarity, attribute preservation, and fast inference through explicit supervision via Triplet ID Group data and an improved diffusion model architecture. Extensive experiments demonstrate its superior performance over existing methods. DreamID represents a significant advancement in face swapping, sets a new paradigm that is remarkably simple and effective, offering high-quality results with rapid inference speed.

{
    \small
    \bibliographystyle{iccv}
    \bibliography{iccv}
}
\clearpage
\setcounter{page}{1}
\maketitlesupplementary
\appendix

\begin{figure*}[htbp] 
\centering 
\includegraphics[width=1.0\textwidth]{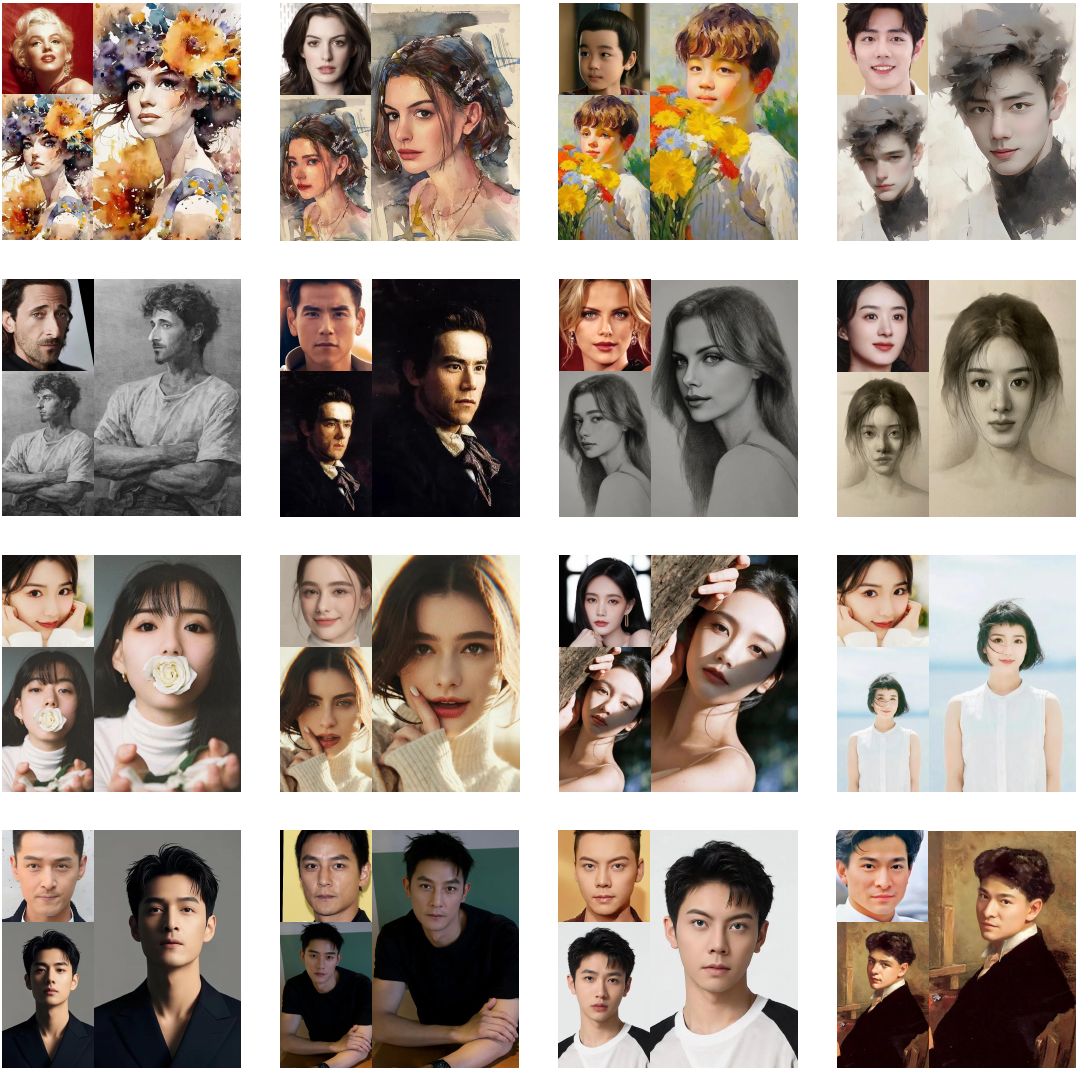} 
\caption{More results of DreamID-High Attribute Preservation.}
\label{fig: more_results_on_image}
\end{figure*}

\section{Ablations about diffusion steps}

We trained two DreamIDs with different inference steps one and four, which are properties inherited from the SD Turbo\cite{sdturbo}. As shown in Table~\ref{Tab:one_four_step_compare}, we tested the qualitative indicators of both the 1 and 4 step models. We found that the performance difference between them was not significant, but the inference time of the 1-step model was substantially shorter. Therefore, we ultimately chose to use the 1-step model.

\begin{table}[htbp]
\centering
\resizebox{\columnwidth}{!}{
\begin{tabular}{ccccccc}
\hline
Model       & FID$\downarrow$   & \begin{tabular}[c]{@{}c@{}}ID \\ Similarity$\uparrow$\end{tabular} & \begin{tabular}[c]{@{}c@{}}ID Retrieval\\ (Top-1/Top-5)$\uparrow$\end{tabular} & Pose$\downarrow$ & Expression$\downarrow$ & \begin{tabular}[c]{@{}c@{}}Inference \\ Time$\downarrow$\end{tabular}\\ \hline
DreamID-4step & 4.69 & 0.71          & 99.9\%/100\%                                  & 2.20 & 0.789 & 1.3s     \\ \hline
DreamID-1step & 5.08 & 0.71          & 99.9\%/100\%                                  & 2.31 & 0.790 & 0.6s \\ \hline
\end{tabular}
}
\caption{Comparison of one step and four step model.}
\label{Tab:one_four_step_compare}
\end{table}

\section{More Results on Image}
We show more image results on Figure~\ref{fig: more_results_on_image}. Our model performs very well in stylized scenes, such as oil paintings, watercolors, sketches, etc. It also handles occlusion effectively, as demonstrated by the flowers in the third row. Additionally, the model excels in preserving light and shadow details. These are all things that previous models could not do, whether it is GAN-based~\cite{e4s,infoswap,inswapper,simswap,simswapplusplus,facedancer,ReliableSwap} or diffusion-based~\cite{ReFace,FaceAdapter,diffswap,DiffFace} model.

\section{DreamID Family}
\begin{figure*}[htbp] 
\centering 
\includegraphics[width=1.0\textwidth]{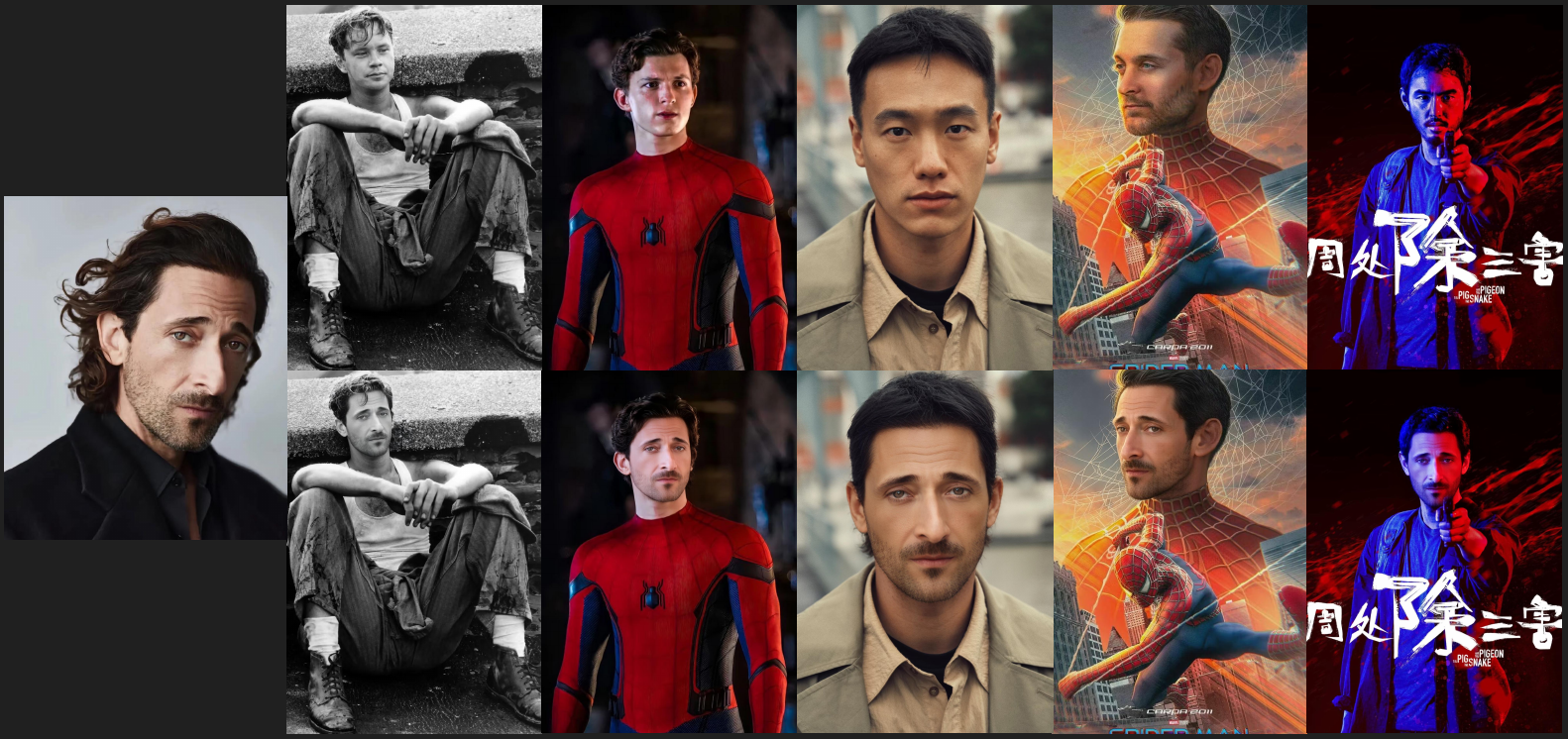} 
\caption{More results of DreamID-High Similarity.}
\label{fig: p1}
\end{figure*}

\begin{figure*}[htbp] 
\centering 
\includegraphics[width=1.0\textwidth]{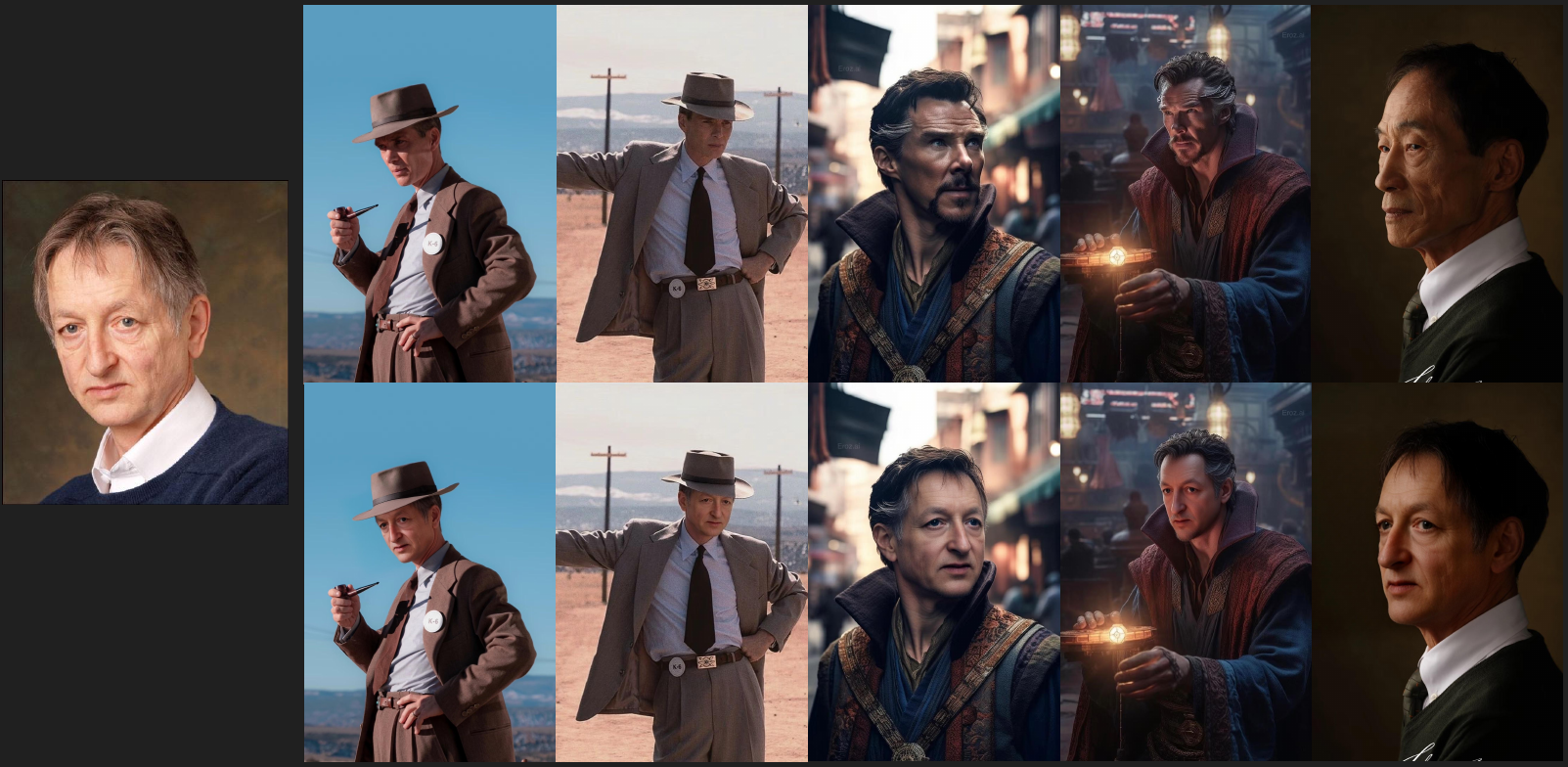} 
\caption{More results of DreamID-High Similarity.}
\label{fig: p2}
\end{figure*}

\begin{figure*}[htbp] 
\centering 
\includegraphics[width=1.0\textwidth]{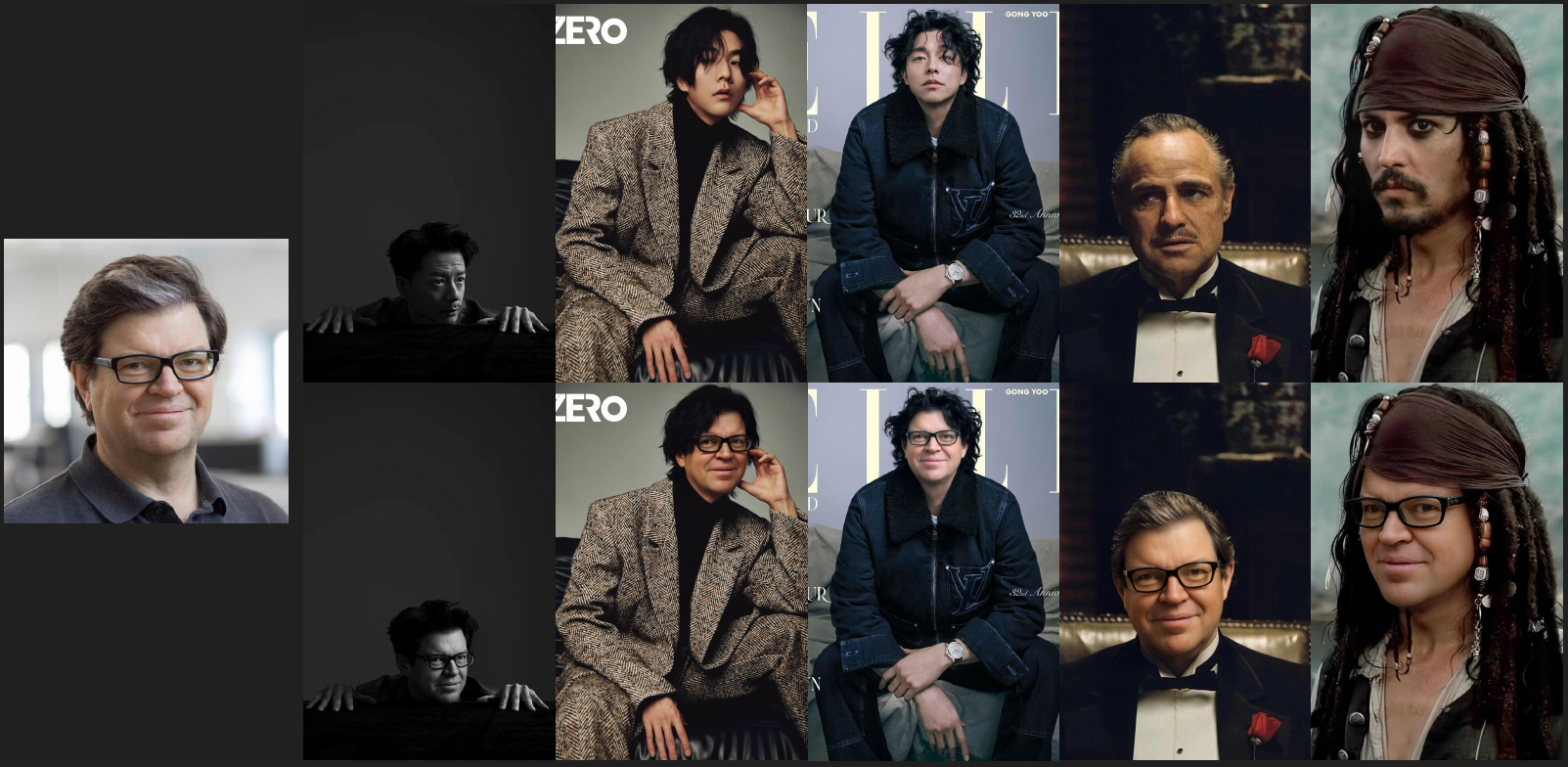} 
\caption{More results of DreamID-High Similarity.}
\label{fig: p3}
\end{figure*}

\begin{figure*}[htbp] 
\centering 
\includegraphics[width=1.0\textwidth]{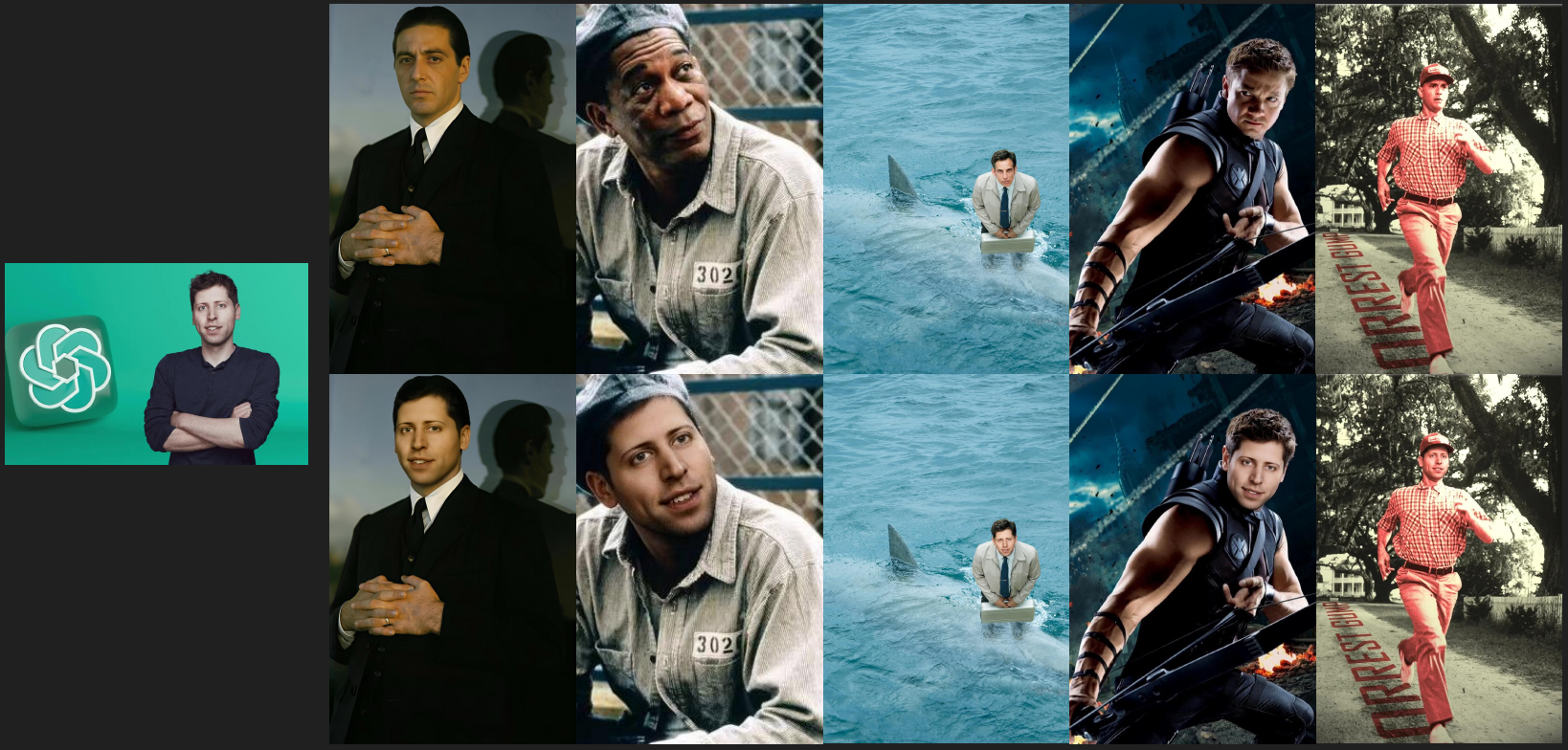} 
\caption{More results of DreamID-High Similarity.}
\label{fig: p4}
\end{figure*}

\begin{figure*}[htbp] 
\centering 
\includegraphics[width=1.0\textwidth]{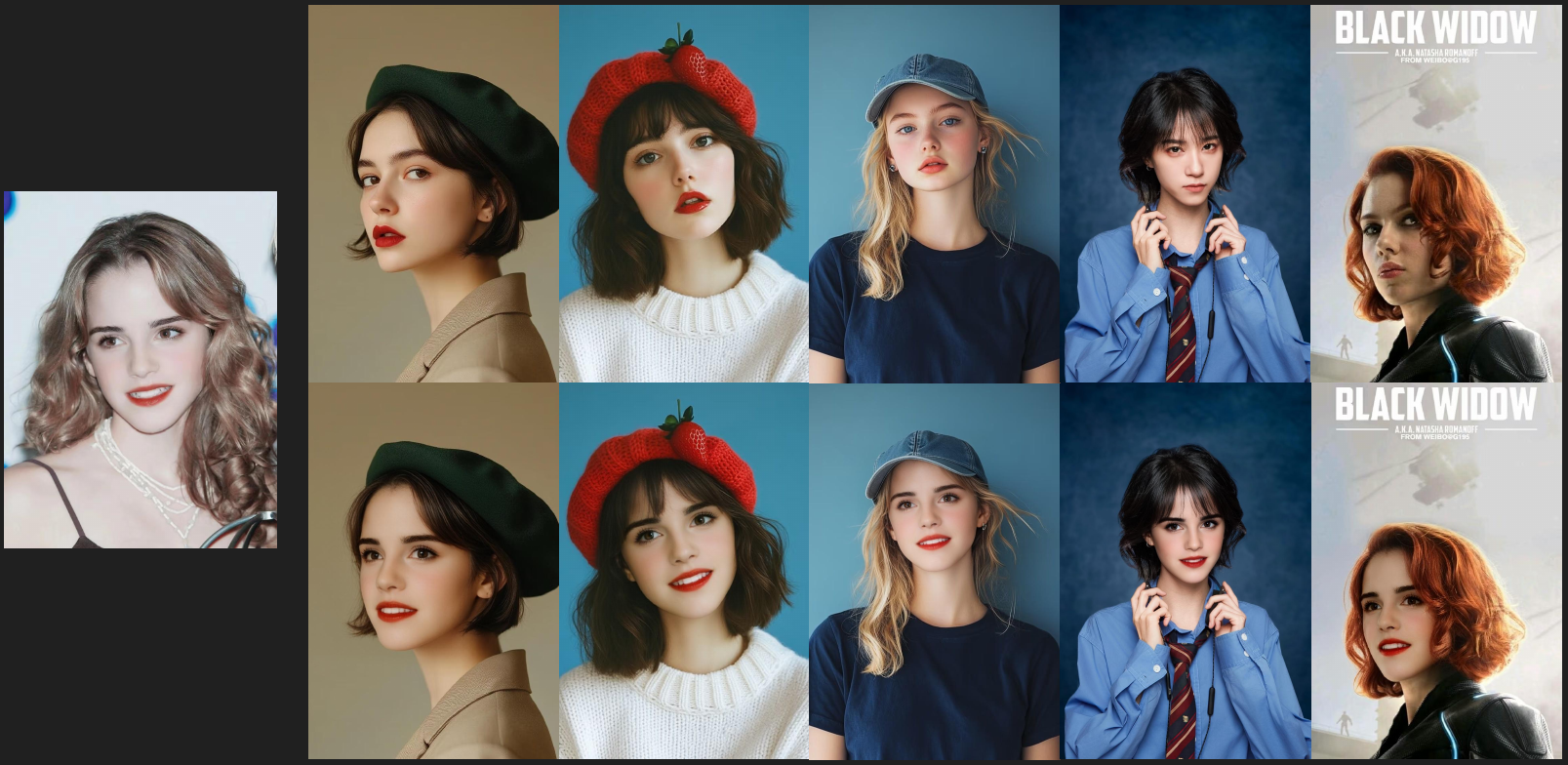} 
\caption{More results of DreamID-High Similarity.}
\label{fig: p5}
\end{figure*}

\begin{figure*}[htbp] 
\centering 
\includegraphics[width=1.0\textwidth]{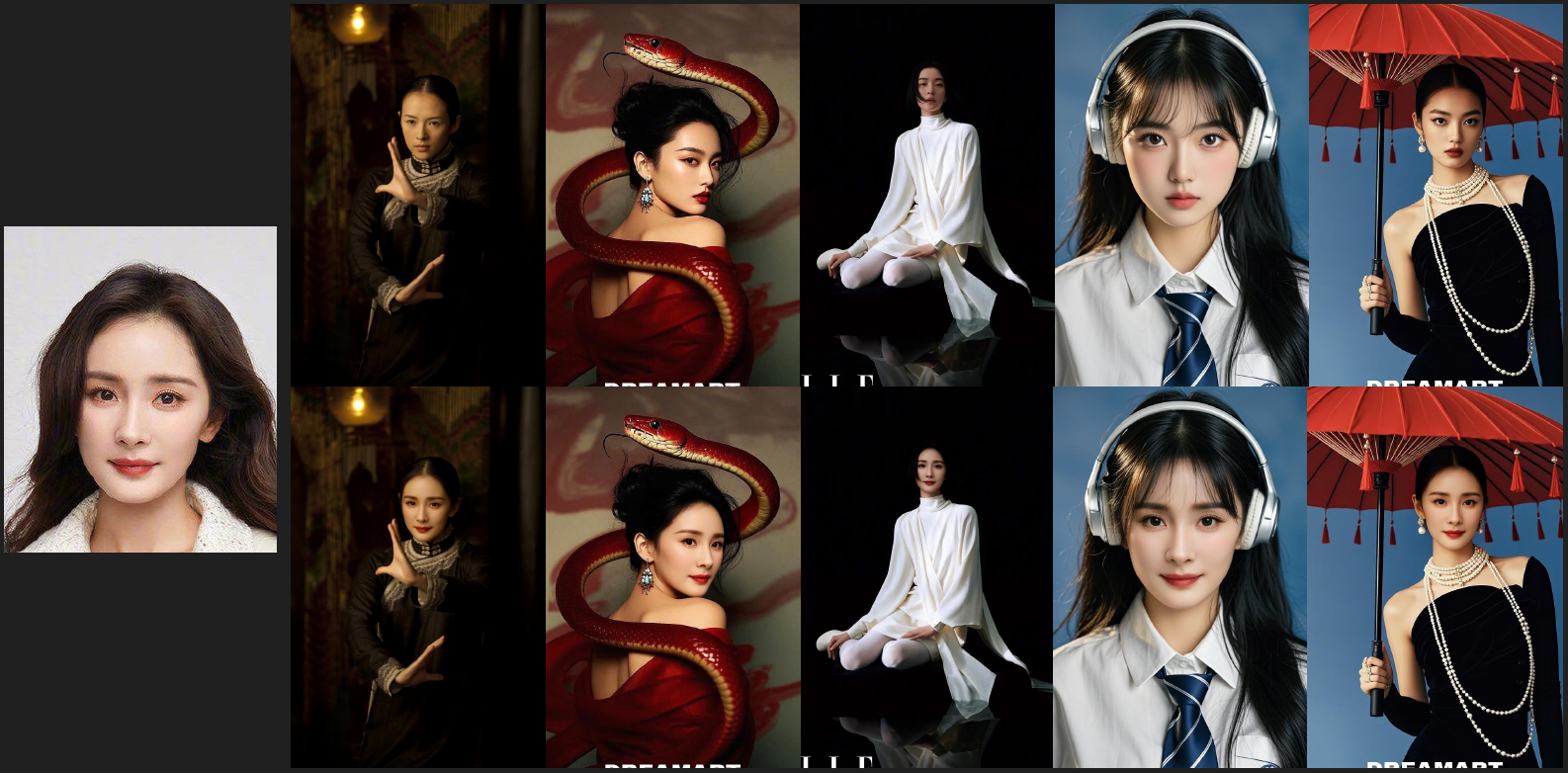} 
\caption{More results of DreamID-High Similarity.}
\label{fig: p6}
\end{figure*}

\begin{figure*}[htbp] 
\centering 
\includegraphics[width=1.0\textwidth]{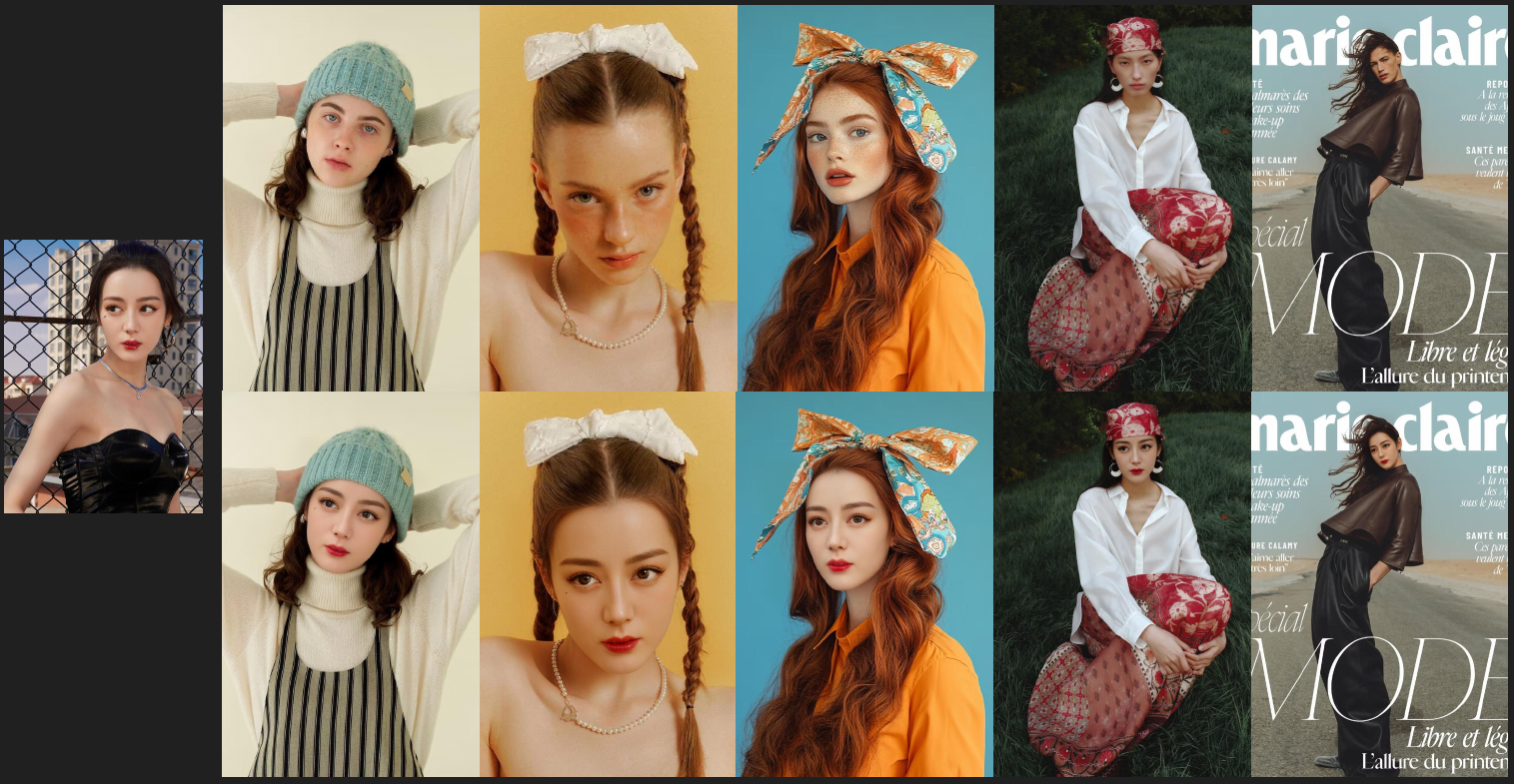} 
\caption{More results of DreamID-High Similarity.}
\label{fig: p7}
\end{figure*}

\begin{figure*}[htbp] 
\centering 
\includegraphics[width=1.0\textwidth]{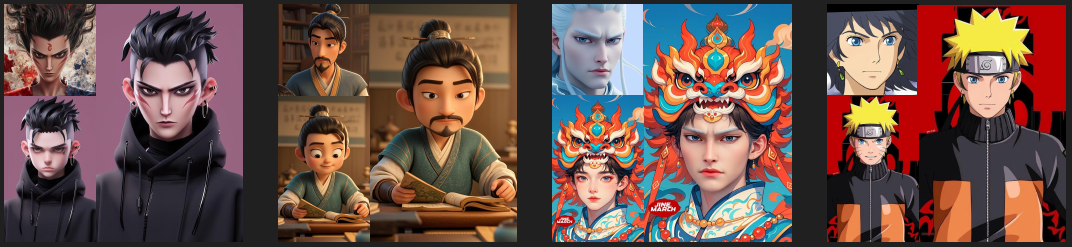} 
\caption{Results of DreamID-Style. DreamID can even perform quite well on stylized user images and stylized target images.}
\label{fig: style_1}
\end{figure*}

\begin{figure*}[htbp] 
\centering 
\includegraphics[width=1.0\textwidth]{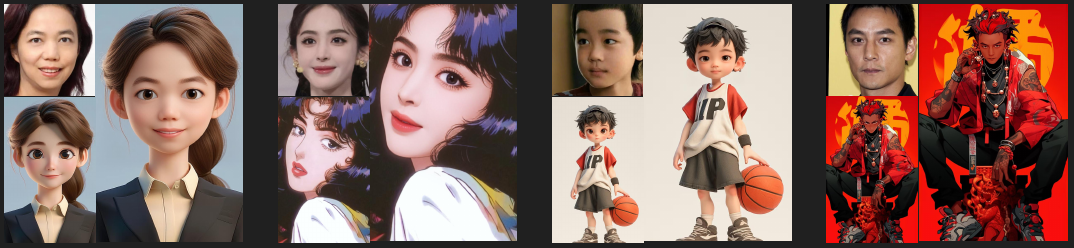} 
\caption{Results of DreamID-Style. DreamID can even perform well on stylized target images, such as 3D and cartoons. This was something that past models were unable to achieve.}
\label{fig: style_2}
\end{figure*}

Our \textbf{Triplet ID Group} confers high flexibility on the face swapping task. By flexibly constructing these triplets, we can train models with different characteristics. Based on this, we have trained several models, including DreamID-High Similarity, DreamID-High Attribute Preservation, and DreamID-Stylization. Figure~\ref{fig: more_results_on_image} show the characteristic of DreamID. DreamID-High Similarity is capable of generating extremely high similarity results, overcoming the problem that traditional face-swapping models are unable to achieve face transformation. DreamID-High Attribute Preservation can effectively preserve fine-grained attribute information, such as lighting/cosmetics, and performs well in handling large angles and occlusions. We show more results of DreamID-High Similarity on Figure~\ref{fig: p1}-\ref{fig: p7}. Previous face swapping work has primarily focused on real human images, with few models supporting stylization such as 3D or cartoon images. We construct stylized triplet ID data by utilizing Pulid\cite{pulid} to train DreamID-Stylization. As shown on Figure~\ref{fig: style_1}, DreamID-Stylization perform well on stylized target images, such as 3D and cartoons. As shown on Figure~\ref{fig: style_2}, DreamID can even perform quite well on stylized user images and stylized target images.

\begin{figure*}[htbp] 
\centering 
\includegraphics[width=1.0\textwidth]{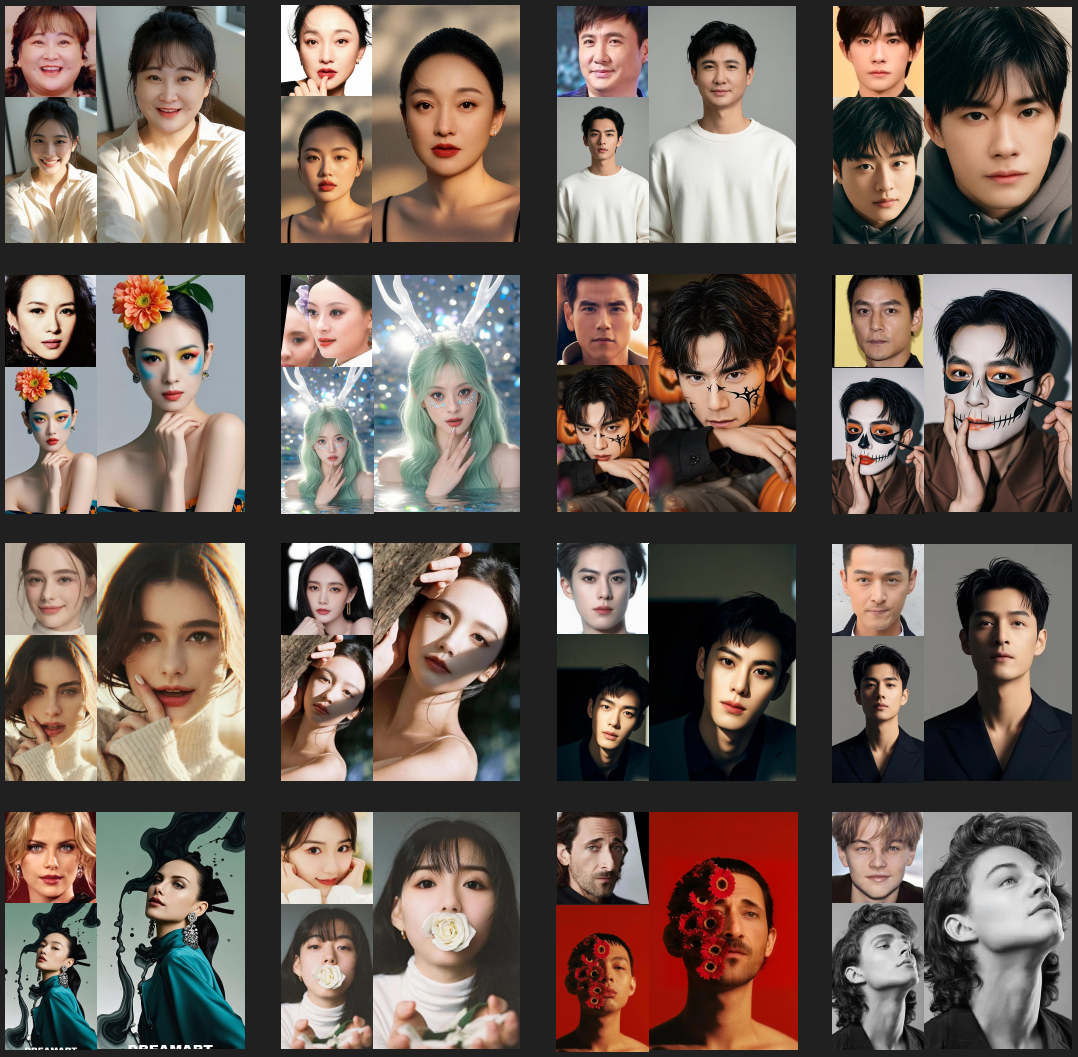} 
\caption{The characteristic of DreamID. DreamID-High Similarity is capable of generating extremely high similarity results, overcoming the problem that traditional face-swapping models are unable to achieve face transformation. DreamID-High Attribute Preservation can effectively preserve fine-grained attribute information, such as lighting/cosmetics, and performs well in handling large angles and occlusions.}
\label{fig: characteristic_of_DreamID}
\end{figure*}

\end{document}